\documentclass[10pt,twocolumn,letterpaper]{article}

\usepackage[pagenumbers]{cvpr} 

\definecolor{cvprblue}{rgb}{0.21,0.49,0.74}
\usepackage[pagebackref,breaklinks,colorlinks,allcolors=cvprblue]{hyperref}
\usepackage{array}      
\usepackage{booktabs}   
\usepackage{multirow}   
\usepackage{amsmath}    
\usepackage{xcolor}     
\usepackage{colortbl}   

\usepackage{caption}    
\usepackage{makecell}
\usepackage{pifont}
\newcommand{\cmark}{\ding{51}}
\newcommand{\xmark}{\ding{55}}
\definecolor{spatialgroup}{HTML}{F0F0F0}
\definecolor{spatialours}{HTML}{DCEBFA}

\def\confName{CVPR}
\def\confYear{2027}

\title{Robo3D-VLM: Grounding Embodied Reasoning in 3D for Robotic Manipulation}

\author{
  Mingchao Qi\textsuperscript{1,*}, \quad
  Zihao Liu\textsuperscript{1,*}, \quad
  Yuanjin Li\textsuperscript{1}, \quad
  Jie Li\textsuperscript{2},\\
  Yuebo Meng\textsuperscript{2}, \quad
  Xing Liu\textsuperscript{1}, \quad
  Zhengxiong Liu\textsuperscript{1}, \quad
  Panfeng Huang\textsuperscript{1}\\[0.25em]
  {\normalsize \textsuperscript{1}Northwestern Polytechnical University \qquad \textsuperscript{2}Xi'an University of Architecture and Technology}\\[0.15em]
  {\small \textsuperscript{*}Equal contribution.}
}

\makeatletter
\patchcmd{\@maketitle}{\vskip .375in}{\vskip .15in}{}{}
\patchcmd{\@maketitle}{\vspace*{24pt}}{\vspace*{10pt}}{}{}
\patchcmd{\@maketitle}{\vspace*{12pt}}{\vspace*{4pt}}{}{}
\makeatother

\begin{document}
\maketitle
\begin{abstract}
Generalizable robotic manipulation requires embodied reasoning: understanding the
scene, identifying what to manipulate, where to make contact, where to place it, and
how it should move. However, existing vision-language models (VLMs) either reason in
the image plane or address only isolated 3D capabilities, leaving a pixel-to-space gap
between reasoning and physical execution. We propose Robo3D-VLM, a 7B VLM that grounds
the full chain of embodied reasoning in 3D. It perceives metric geometry by fusing
pixel-aligned point maps into its visual tokens and expresses every step as 3D points.
Robo3D-VLM first acquires 3D spatial understanding via supervised fine-tuning (SFT),
then extends it to the full chain via reinforcement fine-tuning (RFT) with rewards
computed in 3D. To support training and evaluation, we construct Robo3D-3M, a
large-scale dataset for 3D embodied reasoning, and Robo3D-Bench, a benchmark verifying
it directly in 3D. Robo3D-VLM achieves the best overall performance across six spatial
reasoning benchmarks and on Robo3D-Bench, and reaches 88.4\% and 78.2\% success rates
on SimplerEnv WidowX and Google Robot and 88.0\% across five real-world Franka tasks
without any robot-specific action training, a 38.0-point improvement over the
strongest baseline. It further handles long-horizon and contact-rich tasks with
high-level planners and low-level controllers.
\end{abstract}    
\section{Introduction}
\label{sec:intro}

Embodied reasoning is emerging as a critical pathway toward generalizable robotic
manipulation~\cite{geminiroboticser,embodiedr1}: a robot must understand the 3D
scene well enough to identify the object to manipulate, where to make contact,
where to place it, and how it should move. Vision-language-action (VLA)
models~\cite{rt2,openvla,pi0} imitate expert demonstrations, but the scarcity and
heterogeneity of robot data limit their generalization to unseen scenes. A growing
body of work therefore lets a vision-language model (VLM) reason explicitly and
output an embodiment-agnostic intermediate representation, such as points,
regions, or visual traces, which a downstream planner turns into
motion~\cite{robopoint,fsd,hamster,embodiedr1}. This paradigm enables zero-shot
manipulation without robot-specific action training, while avoiding the error
accumulation of multi-model pipelines in modular approaches~\cite{moka,sofar}.

However, robotic manipulation ultimately takes place in metric 3D space, and
existing methods leave a \emph{pixel-to-space gap} between this requirement and
what VLMs actually produce. Most embodied reasoning models reason in the image
plane: they predict 2D pixel
points~\cite{robopoint,robobrain,embodiedr1,embodiedr15} and recover 3D coordinates
by looking up the depth at those pixels. Since a pixel determines only a viewing
ray, this lookup fails for free-space waypoints, occluded targets, and end-effector
motion. Although 3D-aware VLMs are increasingly common, the embodied reasoning
signals they provide are frequently incomplete and overly simple for complex
decisions, which restricts generalization; most offer only a single, narrow type of
guidance, such as affordance points~\cite{robobrain}, object visual
traces~\cite{fsd}, or target regions~\cite{robopoint}. Moreover, current VLMs
largely acquire such reasoning implicitly through supervised fine-tuning (SFT),
which tends to memorize answers instead of reasoning explicitly, undermining both
generalization and accuracy in open-world 3D embodied reasoning.

\begin{figure*}[t]
  \centering
  \includegraphics[width=0.95\linewidth]{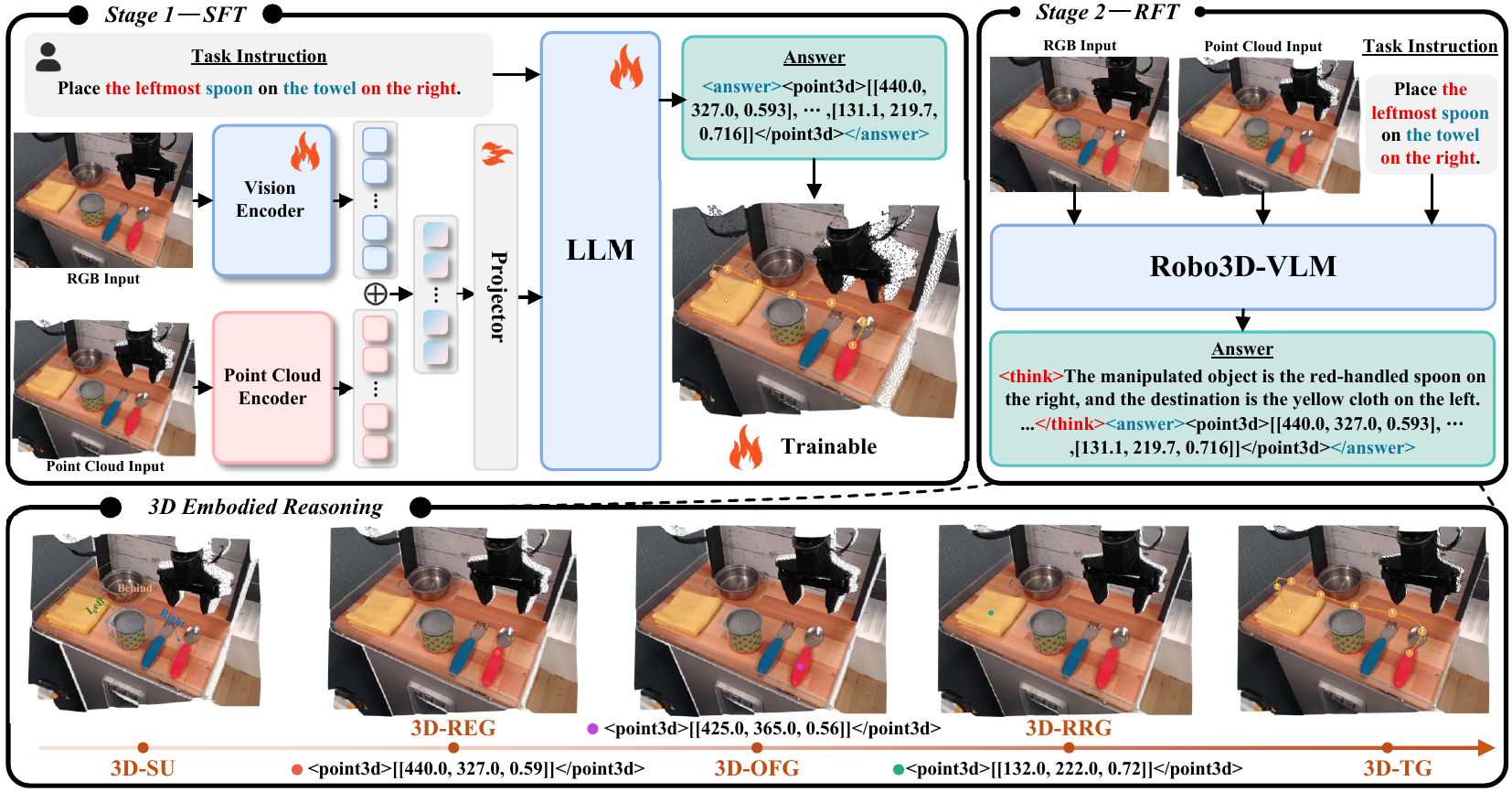}
  \caption{\textbf{Overview of Robo3D-VLM.} Given an RGB image, its aligned point
  cloud, and a task instruction, Robo3D-VLM predicts 3D points $(u,v,d)$ in a unified
  format. In the SFT stage, geometry from the point cloud is added to the
  pixel-aligned visual tokens, equipping the model with 3D spatial understanding.
  In the RFT stage, the model reasons before answering and is optimized with
  modular rewards computed in 3D. The bottom row illustrates the resulting 3D
  embodied reasoning chain on the same scene, from 3D-SU, 3D-REG, 3D-OFG, and
  3D-RRG to 3D-TG.}
  \label{fig:overview}
\end{figure*}

To bridge this gap, we propose Robo3D-VLM, a 7B 3D-aware VLM that not only
acquires 3D spatial understanding via SFT, but also exhibits generalized
reasoning capabilities for the full 3D embodied reasoning chain via RFT. The chain
spans object referring, functional affordance, placement, and object-centric and
end-effector trajectories. Specifically, for 3D spatial understanding,
Robo3D-VLM employs a pixel-aligned 3D visual encoding, which adds an embedding of
the 3D location observed by each image patch to the corresponding visual token,
enhancing metric perception without interfering with the pretrained visual
pathway. To extend this understanding to embodied reasoning, we design an RFT
stage after SFT, in which every step of the chain is expressed as 3D points that
can be directly executed by a motion planner. Since embodied targets in 3D are
rarely unique, SFT tends to imitate a single demonstrated answer, whereas RFT can
reinforce any valid one. We thus introduce a modular reward system (\eg, hit,
distance, and trace) in which every reward is computed in 3D, verifying whether
each prediction satisfies the 3D constraint of its task. To the best of our knowledge, Robo3D-VLM is the first VLM that grounds the full chain of embodied
reasoning in 3D.

To support Robo3D-VLM's training, we present Robo3D-3M, a large-scale dataset with
over 3M samples to first learn 3D spatial understanding in SFT, and further achieve
the full 3D embodied reasoning chain with 260K problems in RFT. It spans general,
indoor, and tabletop scenes with metric 3D annotations (\eg, 3D bounding boxes,
visible functional surfaces, and reference trajectories), lifted from existing 2D,
3D, and robot datasets by fully automated pipelines, including tabletop-level 3D
object annotations reconstructed from robot-view images. Moreover, it contains both
object-centric and end-effector trajectories, enabling complex manipulation such
as operating articulated objects.

We evaluate Robo3D-VLM on six spatial reasoning benchmarks, achieving the best
average rank and outperforming GPT-5.6 Sol and Embodied-R1.5, as well as on four
2D embodied pointing benchmarks. To address the lack of benchmarks for embodied
reasoning in 3D, we introduce Robo3D-Bench with 1,400 test questions verified
directly in 3D, where Robo3D-VLM reaches 72.0\% while no baseline exceeds 9.1\%.
Moreover, Robo3D-VLM achieves 88.4\% and 78.2\% success rates on SimplerEnv WidowX
and Google Robot and 88.0\% across five real-world Franka tasks without any
robot-specific action training, and further executes long-horizon and contact-rich
tasks.

Our contributions are summarized as follows:
\begin{itemize}[leftmargin=*,itemsep=0pt,topsep=2pt]
  \item We propose \textit{Robo3D-VLM}, a 3D-aware VLM that encodes pixel-aligned geometry, predicts 3D points, and is guided by rewards computed in 3D to achieve 3D embodied reasoning.
  \item We construct \textit{Robo3D-3M}, a large-scale dataset tailored for 3D embodied reasoning, and \textit{Robo3D-Bench}, a benchmark that fills the gap in evaluating it.
  \item Extensive experiments show that \textit{Robo3D-VLM} surpasses baselines in 3D spatial understanding and embodied reasoning, and achieves zero-shot manipulation in simulation and the real world.
\end{itemize}
\section{Related Work}
\label{sec:related}

\noindent\textbf{3D Spatial Understanding with VLMs.} Spatial understanding
requires VLMs to perceive object-centric properties (\eg, position, size) and
inter-object relations (\eg, left/right, near/far) in the 3D world. Since VLMs
pre-trained on 2D image-text data lack 3D perception, recent works inject geometry
either explicitly or implicitly. Explicit methods feed depth maps through a shared
image encoder~\cite{spatialbot,spatialrgpt}, a dedicated depth
encoder~\cite{roborefer,robotracer}, or encode 3D coordinates into visual
tokens~\cite{llava3d,video3dllm,n3dvlm}, while implicit methods distill 3D priors
from geometry foundation models~\cite{vggt,spatialmllm,vlm3r}. Despite steady
progress, these models mainly target spatial question answering and 3D grounding,
rather than the embodied reasoning needed to turn 3D understanding into
manipulation.

\noindent\textbf{Intermediate Representations for Robotic Manipulation.}
Intermediate representations abstract away embodiment-specific details and have
become a promising paradigm for generalizable manipulation. Prior works have
explored various forms, including keypoints~\cite{moka,rekep}, affordance
points~\cite{robopoint,robobrain}, object orientations~\cite{sofar}, and visual
traces~\cite{fsd,hamster,magma}, and recent embodied
VLMs~\cite{embodiedr1,embodiedr15,geminiroboticser} unify several of these forms
within a single model. These representations are mostly predicted in 2D and are
used in two ways: (1) lifted to 3D with depth maps and camera intrinsics for
motion planning~\cite{robopoint,embodiedr1}, or (2) rendered onto images to
condition low-level policies~\cite{hamster}. Different from these works, which
mostly operate in 2D or target a single form, Robo3D-VLM covers the full embodied
reasoning chain with a unified 3D representation, including both object-centric
and end-effector trajectories.

\noindent\textbf{Reinforcement Fine-tuning for VLMs.} Reinforcement fine-tuning
(RFT) with algorithms such as GRPO~\cite{grpo} has emerged as a pivotal approach to
enhancing the reasoning of large models~\cite{deepseekr1}. In the multimodal
domain, it has been extended to visual reasoning~\cite{visionr1},
grounding~\cite{visualrft,vlmr1}, and segmentation~\cite{segzero}, and further to
embodied pointing and tracing~\cite{embodiedr1,roborefer,robotracer}. However,
rewards in embodied RFT are mostly computed in the image plane or designed for a
single type of output. We extend RFT to the full 3D embodied reasoning chain with
geometry-verifiable rewards computed directly in 3D.
\section{Method}
\subsection{Problem Formulation}
\label{sec:problem}

We formulate 3D embodied reasoning as predicting 3D points from an RGB image $I$, its
aligned point cloud $P$, and an instruction $L$, where each point $p=(u,v,d)$ comprises
image-plane coordinates $(u,v)$ and absolute depth $d$. The chain comprises five
capabilities sharing this representation. \ding{182}~\textbf{3D Spatial Understanding
(3D-SU)} localizes objects as 3D bounding boxes, each parameterized by a center point
$p$ and its metric size, and reasons about their direction, distance, and size.
\ding{183}~\textbf{3D Referring Expression Grounding (3D-REG)} localizes an object from a
linguistic description by generating 3D points within its oriented 3D bounding box.
\ding{184}~\textbf{3D Object Functional Grounding (3D-OFG)} marks 3D points on the
visible surface of functional parts, such as a handle for grasping.
\ding{185}~\textbf{3D Region Referring Grounding (3D-RRG)} generates 3D points at a
placement region described by relational language (\eg, ``the free space behind the
bowl''). \ding{186}~\textbf{3D Trace Generation (3D-TG)} produces $T$ ordered 3D
waypoints $\tau=\{p_t\}_{t=1}^{T}$ for both object-centric and end-effector
trajectories. These capabilities are tightly coupled: for ``grasp the mug by its
handle and place it behind the bowl'', the affordance points determine where the motion
begins and the placement points where it ends (Fig.~\ref{fig:overview}).

Instead of predicting $(x,y,z)$ directly, we adopt a decoupled $(u,v,d)$ formulation,
convertible to camera-frame coordinates via the intrinsics $K$ as
$\mathbf{x}=d\,K^{-1}[u,v,1]^\top$. This spares the VLM from implicitly learning camera
geometry and aligns its outputs with the pixel-aligned visual tokens, thereby
simplifying training and improving accuracy. Since $d$ is predicted rather than read
from the input depth, waypoints can lie in free space.

\subsection{Robo3D-VLM}
\label{sec:model}

\begin{figure*}[t]
  \centering
  \includegraphics[width=0.95\linewidth]{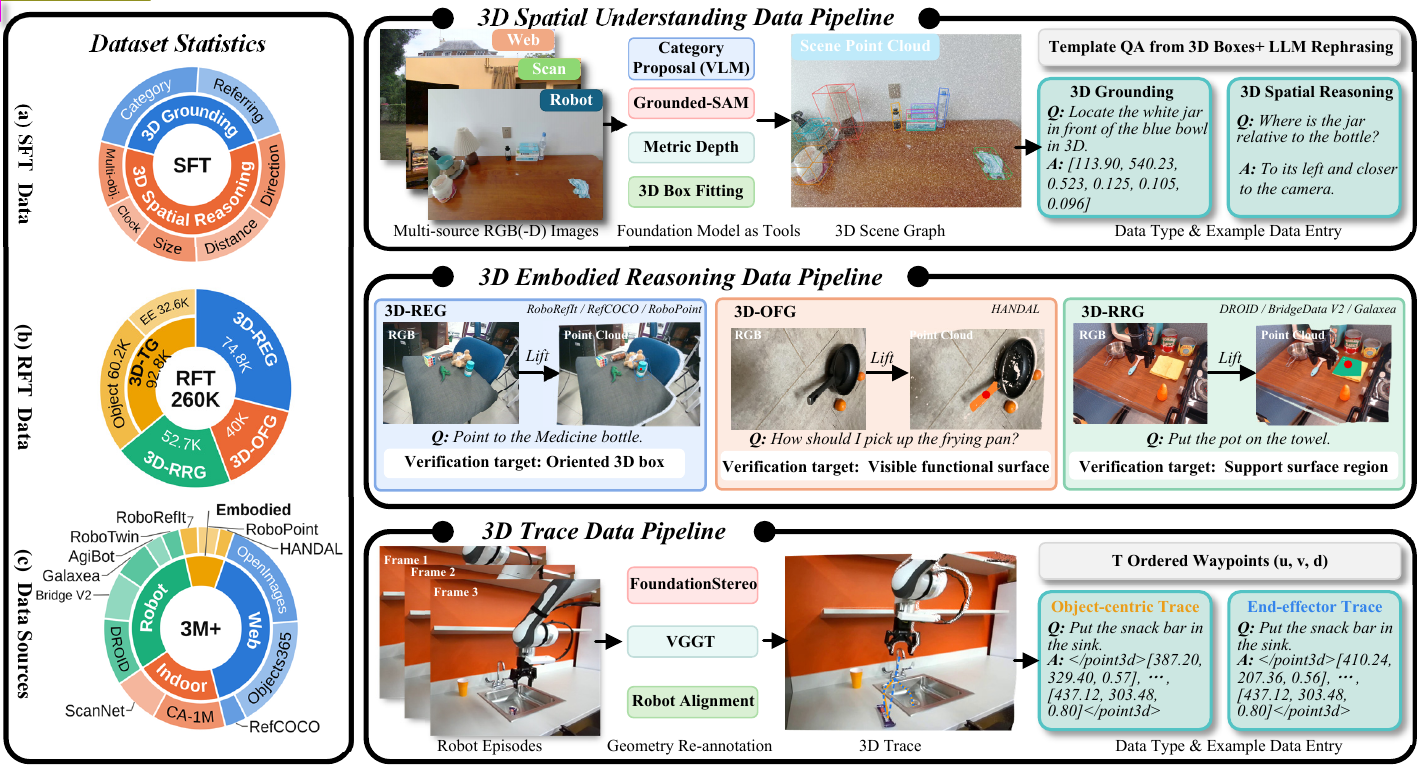}
  \caption{\textbf{Overview of Robo3D-3M.} Left: statistics of (a) SFT data, (b) RFT
  data, and (c) data sources. Right: pipelines for 3D spatial understanding data,
  3D embodied reasoning data, and 3D trajectory data. Each RFT problem is paired with a
  3D verification target used only by the reward.}
  \label{fig:data}
\end{figure*}

\noindent\textbf{VLM Architecture.} Grounding the full chain of embodied reasoning in
3D requires geometry to be available exactly where semantics are. However, 2D VLMs
pretrained solely on RGB images cannot accurately perceive metric geometry, and
existing geometry injection is misaligned: feeding depth through the shared RGB
encoder~\cite{spatialbot,spatialrgpt} interferes with the pretrained image features,
while a separate depth encoder~\cite{roborefer,robotracer} keeps geometry apart from the
visual tokens it describes. Inspired by prior works~\cite{n3dvlm,video3dllm,llava3d},
we introduce a pixel-aligned 3D visual encoding (Fig.~\ref{fig:overview}) that injects 3D
coordinates directly into the visual tokens they belong to, without architectural
changes.
Specifically, given an RGB image $I \in \mathbb{R}^{H\times W\times 3}$ and its aligned
point cloud $P \in \mathbb{R}^{H\times W\times 3}$, the vision encoder extracts image
features $F_{\mathrm{img}} \in \mathbb{R}^{h\times w\times c}$. To align geometry with
these features, we downsample $P$ to $\hat{P} \in \mathbb{R}^{h\times w\times 3}$ by
averaging the valid points within each patch, so that every visual token is associated
with the 3D location of the surface it observes; patches without valid points receive
a zero embedding. Each coordinate $\alpha \in \{x, y, z\}$ of $\hat{P}$ is mapped to a
$c$-dimensional embedding with the standard sinusoidal encoding
$\mathrm{PE}(\cdot)$~\cite{transformer}, and the embeddings of the three axes are summed
and added to the image features:
\begin{equation}
  \tilde{F}_{\mathrm{img}} = F_{\mathrm{img}} + \sum_{\alpha \in \{x,y,z\}} \mathrm{PE}(\alpha).
  \label{eq:pe}
\end{equation}
The fused features are projected by the vision-language merger and fed into the
language model with the instruction tokens. Since geometry enters through addition
rather than as extra tokens, this design preserves the pretrained visual pathway and
leaves the sequence length and inference cost unchanged.

\noindent\textbf{Supervised Fine-tuning.} Since the 2D-only pretraining of general
VLMs limits 3D spatial understanding, Robo3D-VLM first acquires it through a
two-step SFT. (1) \textit{3D Grounding.} We train the model to localize objects
from category names and referring expressions as 3D bounding boxes, teaching it
where objects are and how large they are. (2) \textit{3D Spatial Reasoning.} We
then train it on spatial questions about direction, distance, size, and
multi-object comparison, whose answers are computed from the same object geometry,
mixed with a subset of grounding data to retain localization. All parameters are
updated in both steps. Throughout SFT, we minimize the next-token prediction loss
on the response tokens:
\begin{equation}
  \mathcal{L}_{\mathrm{SFT}} = -\sum_{t} \log \pi_\theta\big(y_t \mid y_{<t}, I, P,
  \mathcal{L}\big),
  \label{eq:sft}
\end{equation}
where $y$ denotes the target response. Moreover, a brief SFT on curated samples of
3D-REG, 3D-OFG, 3D-RRG, and 3D-TG provides a ``cold start'' for the subsequent RFT
stage. See supplementary material for details.

\noindent\textbf{Reinforcement Fine-tuning.} Though SFT builds 3D spatial
understanding, it tends to imitate a single demonstrated answer, and its token-level
loss is agnostic to metric error. However, embodied targets in 3D are rarely unique:
in the example of Sec.~\ref{sec:problem}, any point on the visible handle surface or in
the free space behind the bowl satisfies the instruction. We thus extend Robo3D-VLM
to the full 3D embodied reasoning chain with GRPO~\cite{grpo}, initialized from the
cold-start model. To stabilize multi-task training, we design a modular reward system
in which every reward is computed in 3D, with each predicted point decoded to
camera-frame coordinates as in Sec.~\ref{sec:problem}. \textbf{(1) Format Reward}
($r_{\mathrm{fmt}}$) enforces the \texttt{<think>} tag and the unified output
\texttt{<point3d>[[$u,v,d$],\,...]</point3d>}. \textbf{(2) Hit Reward}
($r_{\mathrm{hit}}$) checks whether a predicted point satisfies the 3D constraint of
its task, \ie, lying inside the oriented 3D box for 3D-REG, on the visible functional
surface for 3D-OFG, and in the target region near the support surface for 3D-RRG.
\textbf{(3) Distance Reward} ($r_{\mathrm{dis}}$) provides dense guidance through a
truncated linear kernel
$\phi(e; e_{\mathrm{lo}}, e_{\mathrm{hi}}) = \mathrm{clip}\big((e_{\mathrm{hi}}-e)/(e_{\mathrm{hi}}-e_{\mathrm{lo}}), 0, 1\big)$,
which grants full credit below $e_{\mathrm{lo}}$ and none beyond $e_{\mathrm{hi}}$, giving
$r_{\mathrm{dis}} = \phi(\delta; \delta_{\min}, \delta_{\max})$, where $\delta$ is the 3D
distance between a predicted point and the target geometry. \textbf{(4) Trace Reward}
for 3D-TG combines $r_{\mathrm{cnt}} = \mathbb{I}(|\hat{\tau}| = T)$ with
$r_{\mathrm{rmse}} = \phi(d_{\mathrm{RMSE}}; D^{R}_{\min}, D^{R}_{\max})$ and
$r_{\mathrm{fr}} = \phi(d_{\mathrm{F}}; D^{F}_{\min}, D^{F}_{\max})$, where
$d_{\mathrm{RMSE}}$ and the discrete Fr\'echet distance $d_{\mathrm{F}}$ measure the
point-wise and order-aware deviation between the predicted trajectory $\hat{\tau}$ and
the reference $\tau$. Each task receives $R = \sum_{r \in \mathcal{F}} w_r r$ with
$\sum_r w_r = 1$, which bounds $R$ within $[0, 1]$ and balances learning across tasks;
responses with an invalid structure receive zero reward. See the supplementary
material for weights and thresholds.

\subsection{Robo3D-3M Dataset}
\label{sec:data}

To equip Robo3D-VLM with the full 3D embodied reasoning chain, we construct
Robo3D-3M, a large-scale dataset with over 3M samples organized into two parts:
3D spatial understanding data for SFT and 3D embodied reasoning data for RFT
(Fig.~\ref{fig:data}). Fully automated pipelines lift the annotations of existing
2D, 3D, and robot datasets into a shared camera-frame 3D space. Since the released
depth and extrinsics of real-robot datasets are often inaccurate, we re-annotate
them by estimating metric depth from stereo recordings with
FoundationStereo~\cite{foundationstereo} and refining VGGT~\cite{vggt} extrinsics
through aligning the rendered robot with the observed depth.

\noindent\textbf{3D Spatial Understanding Data.} This part requires both broad
semantic coverage and fine-grained geometry at the tabletop scale. For general
scenes, we lift OpenImages~\cite{openimages} and Objects365~\cite{objects365} to 3D
with metric point clouds estimated by MoGe-3~\cite{moge3} and completed by
LingBot-Depth~\cite{lingbot}, and fit a 3D box to each back-projected instance
mask. Indoor scans from CA-1M~\cite{ca1m} and ScanNet~\cite{scannet} provide 3D
boxes with precise metric geometry. Since these datasets rarely cover tabletop
manipulation, we further reconstruct 3D object annotations from robot-view images
of Galaxea Open-World~\cite{galaxea}, BridgeData V2~\cite{bridgev2}, and
DROID~\cite{droid} by chaining VLM-based category proposal,
Grounded-SAM~\cite{groundedsam}, the re-annotated depth, and outlier-robust box
fitting. From these 3D boxes, we programmatically generate 3D Grounding QA from
category names or referring expressions, and 3D Spatial Reasoning QA on direction,
distance, size, clock direction, and multi-object comparison, with questions
rephrased by an LLM for diversity.

\noindent\textbf{3D Embodied Reasoning Data.} Since embodied targets in 3D are
rarely unique, each of the 260K problems is paired with 3D verification targets
rather than a single reference answer, which are used only by the reward
(Sec.~\ref{sec:model}). (1) \textbf{3D-REG} provides 74.8K problems from
RoboRefIt~\cite{roborefit}, RefCOCO~\cite{refcoco}, and RoboPoint~\cite{robopoint},
where each target mask is back-projected with the MoGe-3 geometry and fitted with
an oriented 3D box. (2) \textbf{3D-OFG} provides 40K problems from
HANDAL~\cite{handal}, where the meshes of functional parts are rendered into the
current view and filtered by the observed depth to keep only the visible surface.
Questions describe the intended use (\eg, ``Which part should be held to cut
vegetables?''). (3) \textbf{3D-RRG} provides 52.7K problems. For placement, the
final position of the manipulated object in DROID, BridgeData V2, and Galaxea
Open-World is transformed into the initial camera frame and paired with the
support surface from the re-annotated depth. For relational regions, the positive
points of RoboPoint are lifted to 3D. (4) \textbf{3D-TG} provides 92.8K
trajectories, including 60.2K object-centric and 32.6K end-effector traces, from
real episodes in DROID and AgiBot~\cite{agibot} and simulated episodes in
RoboTwin~\cite{robotwin}. After segmenting each episode by gripper-closure events,
we read end-effector traces at the grasp point from the recorded poses. Defined on
the robot rather than the object, they cover both the approach and the grasped
motion, and also specify manipulation whose object motion is hard to describe,
such as opening articulated objects. Object-centric traces are obtained by tracking
the manipulated object with Grounded-SAM, SAM~2~\cite{sam2}, and
Point2Pose~\cite{point2pose}, and refined in GTSAM~\cite{gtsam} with rigid
attachment to the gripper during grasping. Both traces are transformed into the
first camera frame and downsampled to $T$ ordered waypoints. More details are
provided in the supplementary material.
\begin{table*}[t]
  \captionsetup{skip=3pt}
  \caption{Performance comparison on spatial understanding benchmarks. Best results are in
    \textbf{bold} and second best are \underline{underlined}.}
  \label{tab:spatial_understanding}
  \centering
  \footnotesize
  \setlength{\tabcolsep}{2pt}
  \renewcommand{\arraystretch}{0.95}
  \setlength{\aboverulesep}{1pt}
  \setlength{\belowrulesep}{1pt}
  \begin{tabular}{@{}w{l}{80pt}*{14}{w{c}{\dimexpr(\textwidth-80pt-28\tabcolsep)/14\relax}}@{}}
    \toprule
    \multirow{2}{*}{\textbf{Model}}
    & \multicolumn{3}{c}{\textbf{CVBench}}
    & \multicolumn{3}{c}{\textbf{BLINK}}
    & \multicolumn{3}{c}{\textbf{CRPE}}
    & \multicolumn{2}{c}{\textbf{SAT}}
    & \textbf{RoboSp.} & \textbf{EmbSp.} & \multirow{2}{*}{\textbf{Rank}$\boldsymbol{\downarrow}$} \\
    \cmidrule(lr){2-4}\cmidrule(lr){5-7}\cmidrule(lr){8-10}
    \cmidrule(lr){11-12}\cmidrule(lr){13-13}\cmidrule(lr){14-14}
    & \textbf{2DRel} & \textbf{3DDep} & \textbf{3DDis}
    & \textbf{ObjLoc} & \textbf{RelDep} & \textbf{SpRel}
    & \textbf{Subj.} & \textbf{Pred.} & \textbf{Obj.} & \textbf{Val} & \textbf{Real} & \textbf{All} & \textbf{Test} & \\
    \midrule
    GPT-4o & 85.5 & 87.8 & 78.2 & 59.8 & 74.2 & 69.2 & 81.9 & 71.8 & 73.6 & 49.4 & 57.5 & 44.4 & 49.1 & 9.8 \\
    GPT-5.6 Sol & \underline{95.9} & 94.3 & 87.2 & \underline{77.9} & \underline{90.3} & 82.5 & \underline{88.5} & \textbf{76.0} & \textbf{80.6} & 84.0 & \textbf{88.7} & 65.1 & 80.7 & \underline{3.2} \\
    Qwen3-VL-8B & 94.0 & 95.7 & 90.2 & 71.3 & 83.9 & \textbf{86.0} & 77.9 & 68.4 & 74.9 & 55.5 & 74.0 & 67.3 & 79.8 & 5.9 \\
    InternVL3.5-8B & 92.9 & 85.7 & 83.0 & 64.8 & 79.8 & 83.9 & 74.2 & 57.8 & 65.6 & 56.4 & 58.7 & 54.2 & 68.4 & 9.5 \\
    \midrule
    RoboBrain2.5-8B & 94.5 & 96.2 & 91.2 & 61.5 & 86.3 & 82.5 & 75.8 & 65.2 & 70.3 & 59.6 & 67.3 & 63.8 & 75.0 & 7.1 \\
    VeBrain & 84.5 & 77.7 & 76.5 & 51.6 & 71.8 & \underline{85.3} & 78.1 & 62.3 & 70.3 & 60.9 & 66.0 & 44.0 & 67.8 & 9.9 \\
    MiMo-Embodied & \textbf{96.2} & 92.8 & 87.0 & 56.6 & 79.0 & 81.8 & 78.1 & 66.4 & 70.9 & 66.9 & 70.0 & 57.7 & 75.9 & 7.2 \\
    Pelican-VL-1.0 & 86.9 & 82.8 & 81.2 & 51.6 & 79.0 & 83.2 & 79.6 & 62.5 & 69.6 & 59.9 & 66.7 & 56.2 & 70.4 & 9.1 \\
    Embodied-R1.5 & 94.9 & 96.3 & 87.3 & 67.2 & 85.5 & 78.3 & 82.4 & 68.4 & 71.2 & 83.6 & 74.0 & 71.1 & 75.9 & 5.5 \\
    \midrule
    Robo3D-SFT & 94.6 & \underline{96.8} & \underline{91.4} & 76.2 & 89.6 & 80.1 & 87.5 & 72.6 & 78.9 & \underline{84.9} & 76.7 & \underline{78.4} & \underline{86.5} & 3.5 \\
    Robo3D-VLM w/o 3D & 94.8 & 93.1 & 86.2 & 74.9 & 85.3 & 79.8 & 87.9 & 72.9 & 79.1 & 83.2 & 75.8 & 72.6 & 82.1 & 4.9 \\
    \rowcolor{spatialours}
    Robo3D-VLM & 95.2 & \textbf{98.0} & \textbf{93.5} & \textbf{78.7} & \textbf{91.9} & 80.4 & \textbf{89.2} & \underline{74.3} & \underline{80.4} & \textbf{87.1} & \underline{79.3} & \textbf{81.3} & \textbf{89.3} & \textbf{1.9} \\
    \bottomrule
  \end{tabular}
\end{table*} 
\subsection{Action Execution of Robo3D-VLM}
\label{sec:exec}

Since every output of Robo3D-VLM is a set of 3D points in the camera frame, it is
transformed to the robot base frame via the camera extrinsics and executed directly,
without looking up depth or lifting 2D predictions to 3D. The robot thus executes
exactly the geometry the model reasons about, free from background depth errors and able to reach free-space targets. Robo3D-VLM can be coupled with a motion planner for
zero-shot control or, similar to HAMSTER~\cite{hamster}, serve as a high-level
planner for learning-based policies. Here, predicted trajectories are interpolated
and executed with cuRobo~\cite{curobo}. An object-centric trace is mapped to
end-effector waypoints with the grasp offset, whereas an end-effector trace is
followed directly and further specifies how the robot interacts with the object,
which is essential for complex manipulation such as opening articulated objects.
\section{Experiments}
\label{sec:Experiments}

We evaluate Robo3D-VLM in two dimensions: \emph{reasoning}, on six spatial reasoning
benchmarks and the proposed Robo3D-Bench, and \emph{execution}, on seven simulated
tasks in SimplerEnv~\cite{simplerenv} and five real-world tasks on a Franka robot.
All robotic evaluations are conducted zero-shot, without any robot-specific action
training.

\noindent\textbf{Implementation Details.} We adopt
Qwen2.5-VL-7B-Instruct~\cite{qwen25vl} as the base model. First, we perform the
two-step SFT on the 3D spatial understanding data of Robo3D-3M, followed by a brief
cold-start SFT. Then, we continue training with RFT on the 3D embodied reasoning
data of Robo3D-3M. Unless otherwise specified, Robo3D-VLM takes an RGB image and
its aligned point cloud as input, where the point cloud is estimated by
MoGe-3~\cite{moge3} for images without sensor depth. More details are provided in
the supplementary material.

\begin{table}[t]
  \captionsetup{skip=3pt}
  \caption{Performance on embodied pointing benchmarks. W2P, VA-P, and PA denote Where2Place, VABench-Point, and Part-Afford, respectively.}
  \label{tab:pointing}
  \centering
  \footnotesize
  \setlength{\tabcolsep}{1.5pt}
  \renewcommand{\arraystretch}{0.95}
  \setlength{\aboverulesep}{1pt}
  \setlength{\belowrulesep}{1pt}
  \begin{tabular}{@{}w{l}{62pt}*{5}{w{c}{\dimexpr(\columnwidth-62pt-10\tabcolsep)/5\relax}}@{}}
    \toprule
    \textbf{Model} & \textbf{RoboRefIt} & \textbf{W2P} & \textbf{VA-P} & \textbf{PA} & \textbf{Robo3D-P} \\
    \midrule
    GPT-5.6 Sol    & 86.9 & 61.5 & 59.5 & 70.8 & \underline{9.1} \\
    Qwen3-VL-8B    & 83.6 & 64.0 & 42.3 & 48.7 & 3.8 \\
    MiMo-Embodied  & 82.3 & 63.6 & 46.9 & 65.5 & 0.4 \\
    RoboBrain2.5   & 84.6 & 70.0 & 39.0 & 25.2 & 7.3 \\
    Pelican-VL-1.0 & 74.9 & 57.8 & 14.5 & 28.8 & 0.1 \\
    Embodied-R1.5  & \underline{88.4} & \underline{74.0} & \underline{73.7} & \underline{82.9} & 8.5 \\
    \rowcolor{spatialours}
    Robo3D-VLM     & \textbf{89.2} & \textbf{75.0} & \textbf{77.7} & \textbf{87.1} & \textbf{72.0} \\
    \bottomrule
  \end{tabular}
\end{table}

\subsection{Evaluation of 3D Spatial Understanding}
\label{sec:exp_spatial}

\noindent\textbf{Setup.} We evaluate on six public benchmarks: CV-Bench~\cite{cvbench},
BLINK~\cite{blink}, CRPE~\cite{crpe}, SAT~\cite{sat}, RoboSpatial~\cite{robospatial},
and EmbSpatial-Bench~\cite{embspatial}, with point clouds estimated by
MoGe-3~\cite{moge3}. Baselines include general-purpose VLMs (\eg, GPT-5.6 Sol and
Qwen3-VL-8B) and embodied VLMs (\eg, RoboBrain2.5, MiMo-Embodied, and Embodied-R1.5),
and we report the average rank over all 13 metrics. We also report two ablations:
Robo3D-SFT, trained on the same data with SFT in place of RFT, and Robo3D-VLM w/o 3D,
which takes only RGB input.

\noindent\textbf{Results.} As shown in Tab.~\ref{tab:spatial_understanding}, Robo3D-VLM
achieves the best average rank of 1.9 with only 7B parameters, outperforming GPT-5.6
Sol (3.2) and Embodied-R1.5 (5.5). The gains are most pronounced on 3D metrics: it
ranks first on 3D depth and distance in CV-Bench and relative depth in BLINK, and
surpasses the second-best baseline by 10.2 and 8.6 points on RoboSpatial and
EmbSpatial-Bench. Removing the 3D visual encoding drops these five 3D metrics by 6.9
points on average but the 2D relation metrics by only 0.5, indicating that the gains
stem from metric 3D awareness. Moreover, Robo3D-VLM outperforms Robo3D-SFT on all 13
metrics, showing that RFT on the embodied reasoning chain also strengthens 3D spatial
understanding.

\begin{table}[t]
  \captionsetup{skip=3pt}
  \caption{Performance comparison on visual trace benchmarks. ``--'' indicates no support
for 3D trace output. Lowest errors are in \textbf{bold} and second lowest are
\underline{underlined}.}
  \label{tab:trace}
  \centering
  \footnotesize
  \setlength{\tabcolsep}{1.5pt}
  \renewcommand{\arraystretch}{0.95}
  \setlength{\aboverulesep}{1pt}
  \setlength{\belowrulesep}{1pt}
  \begin{tabular}{@{}w{l}{58pt}*{6}{w{c}{\dimexpr(\columnwidth-58pt-12\tabcolsep)/6\relax}}@{}}
    \toprule
    \multirow{2}{*}{\textbf{Model}}
    & \multicolumn{2}{c}{\textbf{ShareRobot-V}}
    & \multicolumn{2}{c}{\textbf{VABench-V}}
    & \multicolumn{2}{c}{\textbf{Robo3D-T}} \\
    \cmidrule(lr){2-3}\cmidrule(lr){4-5}\cmidrule(lr){6-7}
    & RMSE$\downarrow$ & DFD$\downarrow$ & RMSE$\downarrow$ & DFD$\downarrow$ & RMSE$\downarrow$ & DFD$\downarrow$ \\
    \midrule
    GPT-5.6 Sol    & 19.63 & 27.89 & 11.45 & 16.04 & 20.07 & 28.55 \\
    Qwen3-VL-8B    & 22.58 & 32.47 & 15.30 & 21.80 & 39.43 & 47.69 \\
    RoboBrain2.5   & \underline{14.24} & \underline{19.97} & 14.04 & 19.42 & \underline{18.10} & \underline{25.70} \\
    Embodied-R1    & 23.20 & 32.90 & \underline{9.04} & \underline{12.21} & -- & -- \\
    Embodied-R1.5  & 15.79 & 22.64 & \textbf{7.00} & \textbf{9.83} & 21.50 & 30.50 \\
    \rowcolor{spatialours}
    Robo3D-VLM     & \textbf{13.42} & \textbf{18.28} & 9.11 & 12.82 & \textbf{9.65} & \textbf{14.52} \\
    \bottomrule
  \end{tabular}
\end{table}

\begin{table*}[t]
  \captionsetup{skip=3pt}
  \caption{SimplerEnv evaluation on WidowX and Google Robot. Best results are in
\textbf{bold} and second best are \underline{underlined}.}
  \label{tab:simplerenv}
  \centering
  \footnotesize
  \setlength{\tabcolsep}{2pt}
  \setlength{\arrayrulewidth}{0.4pt}
  \renewcommand{\arraystretch}{0.95}
  \setlength{\aboverulesep}{1pt}
  \setlength{\belowrulesep}{1pt}
  \begin{tabular}{@{}w{l}{80pt}|*{5}{w{c}{\dimexpr(\textwidth-80pt-18\tabcolsep-2\arrayrulewidth)/9\relax}}|*{4}{w{c}{\dimexpr(\textwidth-80pt-18\tabcolsep-2\arrayrulewidth)/9\relax}}@{}}
    \toprule
    \multirow[c]{2}{*}{\textbf{Method}}
    & \multicolumn{5}{c|}{\textbf{WidowX Robot (Visual Matching)}}
    & \multicolumn{4}{c}{\textbf{Google Robot (Variant Aggregation)}} \\
    \cmidrule(lr){2-6}\cmidrule(l){7-10}
    & Spoon & Carrot & Stack & Eggplant & Avg.
    & Coke Can & Move Near & Drawer & Avg. \\
    \midrule
    OpenVLA      & 4.2 & 0.0 & 0.0 & 12.5 & 4.2 & 60.8 & 67.7 & 28.8 & 52.4 \\
    SpatialVLA   & 16.7 & 25.0 & 29.2 & \textbf{100.0} & 42.7 & 88.0 & 72.7 & 41.8 & 67.5 \\
    $\pi_0$-FAST & 29.1 & 21.9 & 10.8 & 66.6 & 32.1 & 77.6 & 68.2 & 31.3 & 59.0 \\
    $\pi_{0.5}$  & 49.3 & 64.7 & 44.7 & 69.7 & 57.1 & 87.4 & \underline{73.5} & 50.5 & 70.5 \\
    GR00T-N1.5   & \underline{75.3} & 54.3 & 57.0 & 61.3 & \underline{62.0} & 69.3 & 68.7 & 35.8 & 57.9 \\
    GR00T-N1.6   & 64.5 & 65.5 & 5.5 & \underline{93.0} & 57.1 & 89.6 & 68.3 & \textbf{59.5} & \underline{72.5} \\
    \midrule
    MOKA         & 45.8 & 41.6 & 33.3 & 12.5 & 33.3 & 56.1 & 45.8 & 18.3 & 40.1 \\
    SoFar        & 55.5 & 56.9 & \underline{62.5} & 40.2 & 53.8 & \underline{93.1} & 62.5 & 40.3 & 65.3 \\
    \midrule
    RoboPoint    & 16.7 & 20.8 & 8.3 & 25.0 & 17.7 & 18.1 & 20.8 & 0.0 & 13.0 \\
    FSD          & 41.6 & 50.0 & 33.3 & 37.5 & 40.6 & 87.4 & 53.8 & 0.0 & 47.1 \\
    Embodied-R1  & 62.5 & \underline{68.0} & 36.1 & 58.3 & 56.2 & \underline{93.1} & \textbf{79.2} & 25.0 & 65.8 \\
    \rowcolor{spatialours}
    Robo3D-VLM (Ours) & \textbf{95.8} & \textbf{91.7} & \textbf{65.9} & \textbf{100.0} & \textbf{88.4} & \textbf{97.2} & \textbf{79.2} & \underline{58.3} & \textbf{78.2} \\
    \bottomrule
  \end{tabular}
\end{table*}

\subsection{Evaluation of 3D Embodied Reasoning}
\noindent\textbf{Robo3D-Bench.} Since existing embodied pointing benchmarks evaluate
predictions only in the image plane, we present Robo3D-Bench, a challenging benchmark
that verifies embodied reasoning directly in 3D across diverse real-world and simulated
manipulation scenes. It contains 1,400 test questions in two sub-benchmarks. Robo3D-P
has 800 questions evenly covering object referring, spatial relations, functional
contact surfaces, and placement regions, each answered with a single 3D point that
succeeds only if it lies within 5\,cm of the target geometry and its projection falls
inside the target region. Robo3D-T has 600 trajectories evenly split between
object-centric and end-effector traces, each answered with 8 ordered 3D waypoints and
evaluated by RMSE and DFD. All test questions are disjoint from the training data at
both the image and episode levels.

\noindent\textbf{Setup.} We evaluate 3D-REG on RoboRefIt~\cite{roborefit}, 3D-RRG on
Where2Place~\cite{robopoint} and VABench-Point~\cite{fsd}, and 3D-OFG on
Part-Afford~\cite{embodiedr1} by projecting the predicted 3D points onto the image,
while Robo3D-P evaluates all three capabilities directly in 3D. For 3D-TG, we follow
ShareRobot-V~\cite{robobrain} and VABench-V~\cite{fsd} for 2D traces and use Robo3D-T
for both object-centric and end-effector traces in 3D. On Robo3D-Bench, each baseline
predicts 3D points through its native output interface from the RGB image and camera
intrinsics. Results are presented in Tabs.~\ref{tab:pointing} and~\ref{tab:trace}.

\noindent\textbf{(O1) Existing VLMs suffer from a severe pixel-to-space gap.} Baselines
perform reasonably on 2D pointing benchmarks, \eg, Embodied-R1.5 reaches 88.4\% on
RoboRefIt and 82.9\% on Part-Afford, yet none exceeds 9.1\% on Robo3D-P: even when
their projections fall inside the target region, the predicted depth often deviates
by more than 5\,cm. Accurate pointing in the image plane thus does not translate into
correct grounding in 3D, which cannot be obtained by simply attaching depth to 2D
predictions.

\begin{figure}[t]
  \centering
  \includegraphics[width=\linewidth]{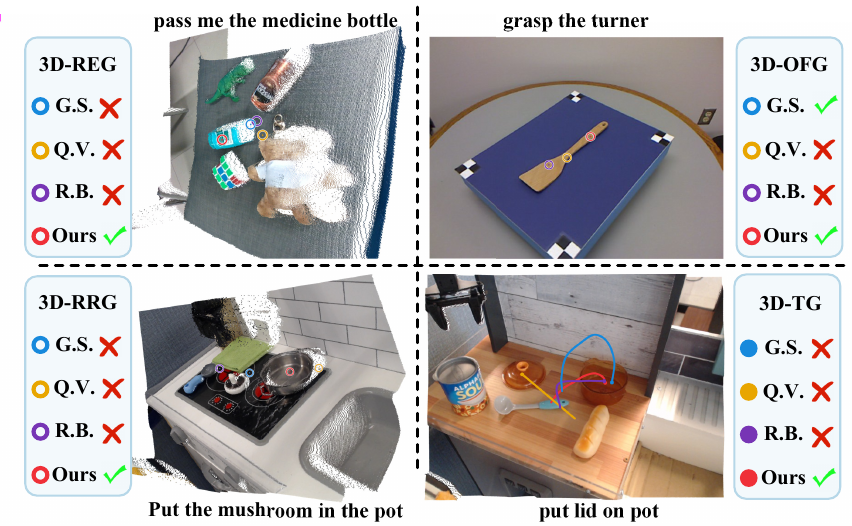}
  \caption{\textbf{Qualitative comparison of 3D embodied reasoning.} We compare
  GPT-5.6 Sol (G.S.), Qwen3-VL-8B (Q.V.), RoboBrain 2.5 (R.B.), and Robo3D-VLM
  (Ours) on Robo3D-Bench. All predictions are visualized in the point cloud.
  \cmark\ and \xmark\ denote success and failure under the 3D criterion.}
  \label{fig:vis_reasoning}
\end{figure}

\noindent\textbf{(O2) Robo3D-VLM grounds every capability of the chain accurately in
3D.} As shown in Tab.~\ref{tab:pointing}, Robo3D-VLM achieves the best results on all
four 2D pointing benchmarks across 3D-REG, 3D-RRG, and 3D-OFG, showing that reasoning
in 3D does not compromise accuracy in the image plane, and reaches 72.0\% on Robo3D-P,
outperforming the strongest baseline by over 60 points. As visualized in
Fig.~\ref{fig:vis_reasoning}, the baselines often place points on the wrong object or
off the target in 3D, whereas Robo3D-VLM succeeds on every step of the chain.

\noindent\textbf{(O3) Robo3D-VLM generates accurate 3D trajectories for both objects
and end-effectors.} As shown in Tab.~\ref{tab:trace}, Robo3D-VLM reduces the RMSE and
DFD of the strongest baseline on Robo3D-T by 47\% and 44\%, while 2D-only models such
as Embodied-R1 cannot produce 3D trajectories at all. It also achieves the lowest
errors on ShareRobot-V and remains competitive on VABench-V.

\subsection{Evaluation of Robo3D-VLM for Robot Manipulation}
\label{sec:exp_manip}
\begin{figure*}[t]
  \centering
  \includegraphics[width=0.85\linewidth]{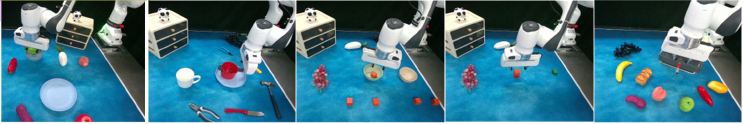}
  \caption{\textbf{Real-world tasks.} From left to right, Apple to Bowl (3D-REG),
  Pot on Plate (3D-OFG), Middle Block (3D-SU), Stack Blocks (3D-RRG), and Open
  Drawer (3D-TG), each probing a different step of the 3D embodied reasoning chain.}
  \label{fig:real_tasks}
\end{figure*}

\noindent\textbf{SimplerEnv Simulation.} We evaluate zero-shot manipulation in
SimplerEnv~\cite{simplerenv} under the Visual Matching setting on WidowX and the
Variant Aggregation setting on Google Robot, which varies backgrounds, lighting,
distractors, table textures, and camera poses, executing the predicted 3D trajectories
as in Sec.~\ref{sec:exec}. Baselines include VLA models (OpenVLA~\cite{openvla}, SpatialVLA~\cite{spatialvla}, $\pi_0$-FAST~\cite{pi0fast},
$\pi_{0.5}$~\cite{pi05}, and GR00T-N1.5/N1.6~\cite{gr00t}), modular methods that
cascade foundation models (MOKA~\cite{moka} and SoFar~\cite{sofar}), and embodied
VLMs that predict affordance points or visual traces (RoboPoint~\cite{robopoint},
FSD~\cite{fsd}, and Embodied-R1~\cite{embodiedr1}). As shown in
Tab.~\ref{tab:simplerenv}, Robo3D-VLM achieves 88.4\% and 78.2\%, surpassing the
strongest baselines by 26.4 and 5.7 points without any robot-specific action
training, and its lead under Variant Aggregation demonstrates robustness to visual
variations. Compared with embodied VLMs that reason in the image plane, the gains
reflect the value of grounding embodied reasoning in 3D. On Stack Blocks, whose target
lies in free space above the bottom block and cannot be recovered from 2D predictions,
Robo3D-VLM improves the best embodied VLM result from 36.1\% to 65.9\%. Open/Close
Drawer involves complex articulated manipulation that cannot be specified by
object-centric traces alone; by predicting 3D end-effector trajectories, Robo3D-VLM
achieves 58.3\%, whereas modular methods and embodied VLMs achieve at most 40.3\%.

\begin{table}[t]
  \captionsetup{skip=3pt}
    \caption{Real-world evaluation on a Franka robot (10 trials per task). VLA models are
    fine-tuned with 50 demonstrations per task, and the other methods are zero-shot.}
  \label{tab:realworld}
  \centering
  \footnotesize
  \setlength{\tabcolsep}{1.5pt}
  \renewcommand{\arraystretch}{0.95}
  \setlength{\aboverulesep}{1pt}
  \setlength{\belowrulesep}{1pt}
  \begin{tabular}{@{}w{l}{70pt}*{6}{w{c}{\dimexpr(\columnwidth-70pt-12\tabcolsep)/6\relax}}@{}}
    \toprule
    \textbf{Method}
    & \makecell[c]{Apple\\to Bowl}
    & \makecell[c]{Pot on\\Plate}
    & \makecell[c]{Middle\\Block}
    & \makecell[c]{Stack\\Blocks}
    & \makecell[c]{Open\\Drawer}
    & \makecell[c]{Avg.} \\
    \midrule
    OpenVLA-OFT  & 80.0 & 40.0 & 10.0 & 40.0 & 10.0 & 36.0 \\
    $\pi_{0.5}$  & 90.0 & 60.0 & 10.0 & 50.0 & 10.0 & 44.0 \\
    MOKA         & 60.0 & 40.0 & 50.0 & 10.0 & 50.0 & 42.0 \\
    Embodied-R1  & \textbf{100.0} & 70.0 & 60.0 & 20.0 & 0.0 & 50.0 \\
    \rowcolor{spatialours}
    Robo3D-VLM (Ours) & \textbf{100.0} & \textbf{100.0} & \textbf{90.0} & \textbf{70.0} & \textbf{80.0} & \textbf{88.0} \\
    \bottomrule
  \end{tabular}
\end{table}
\noindent\textbf{Real-World Robot Evaluation.} We deploy Robo3D-VLM on a Franka robot
with a third-person Intel RealSense D455 RGB-D camera, and design five tasks
(Fig.~\ref{fig:real_tasks}) that probe different steps of the 3D embodied reasoning
chain. Each task is tested over 10 trials in unseen scenes. We compare against
OpenVLA-OFT~\cite{openvlaoft} and $\pi_{0.5}$~\cite{pi05}, fine-tuned with 50
demonstrations per task, as well as MOKA~\cite{moka} and
Embodied-R1~\cite{embodiedr1}, evaluated zero-shot like ours.

As shown in Tab.~\ref{tab:realworld}, Robo3D-VLM achieves an 88.0\% zero-shot success
rate, outperforming the strongest baseline by 38.0 points. The fine-tuned VLAs succeed
in simple pick-and-place but drop to 10.0\% on Middle Block and Open Drawer, as
imitation from limited demonstrations fails to capture the spatial relations in the
instruction. Among zero-shot baselines, MOKA suffers from error accumulation across its
cascaded modules, while Embodied-R1 can only lift its 2D points to the observed surface,
which collapses the stacking target onto the red block (20.0\%), and its object-centric
traces cannot specify how the gripper pulls the drawer (0.0\%). In contrast,
Robo3D-VLM achieves at least 70.0\% on every task, as every step of its reasoning is
grounded in 3D and executed directly.

\begin{figure}[t]
  \centering
  \includegraphics[width=0.85\linewidth]{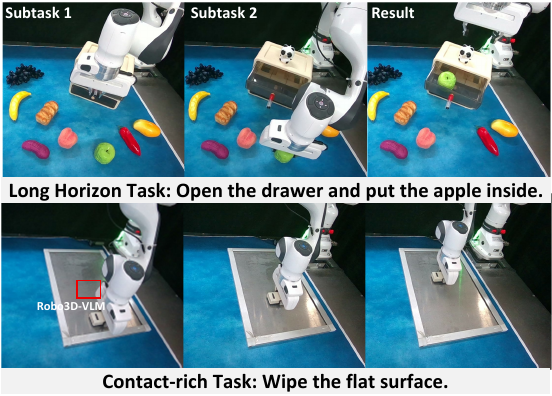}
  \caption{\textbf{Long-horizon and contact-rich manipulation.} In the top row, GPT-4o
  decomposes the instruction into sub-tasks, each solved by Robo3D-VLM in 3D. In the
  bottom row, Robo3D-VLM wipes off the writing in the red box with a predicted 3D
  end-effector trajectory under impedance control.}
  \label{fig:qualitative}
\end{figure}

\noindent\textbf{Qualitative Analysis in Complex Scenarios.} We further examine how the
3D embodied reasoning chain extends beyond single-step manipulation
(Fig.~\ref{fig:qualitative}). (1) Long-horizon tasks. For instructions such as ``open
the drawer and put the green apple into it'', GPT-4o~\cite{gpt4o} decomposes the
instruction into sub-tasks, and Robo3D-VLM solves each on the current observation,
\eg, an end-effector trajectory to pull the drawer open, followed by referring and
placement for the apple inside the revealed drawer. (2) Contact-rich manipulation.
Wiping the writing off a board involves no object to be moved, so its motion can only
be specified on the robot side. Robo3D-VLM predicts a 3D end-effector trajectory along
the board surface, executed with an impedance controller to maintain stable contact.

\begin{table}[t]
  \captionsetup{skip=3pt}
  \caption{Ablation study. Reward denotes whether rewards are computed in 3D or in the image plane (2D).}
  \label{tab:ablation}
  \centering
  \footnotesize
  \setlength{\tabcolsep}{1.5pt}
  \renewcommand{\arraystretch}{0.95}
  \setlength{\aboverulesep}{1pt}
  \setlength{\belowrulesep}{1pt}
  \begin{tabular}{@{}w{c}{9pt}w{c}{18pt}w{c}{19pt}w{c}{22pt}w{c}{27pt}w{c}{24pt}w{c}{36pt}*{2}{w{c}{\dimexpr(\columnwidth-155pt-16\tabcolsep)/2\relax}}@{}}
    \toprule
    \multirow{2}{*}{ID} & 3D & \multirow{2}{*}{RFT} & \multirow{2}{*}{Think}
      & \multirow{2}{*}{Reward} & W2P & Robo3D-P & \multicolumn{2}{c}{Robo3D-T} \\
    \cmidrule(lr){8-9}
      & Enc. & & & & SR$\uparrow$ & SR$\uparrow$ & RMSE$\downarrow$ & DFD$\downarrow$ \\
    \midrule
    A & \xmark & \cmark & \cmark & 3D & 73.8 & 41.5 & 15.82 & 22.47 \\
    B & \cmark & \xmark & \xmark & --  & 63.5 & 45.2 & 13.94 & 20.15 \\
    C & \cmark & \cmark & \xmark & 3D & 73.2 & 67.4 & 10.38 & 15.61 \\
    D & \cmark & \cmark & \cmark & 2D & 74.6 & 52.6 & 12.85 & 18.93 \\
    \rowcolor{spatialours}
    E & \cmark & \cmark & \cmark & 3D & \textbf{75.0} & \textbf{72.0} & \textbf{9.65} & \textbf{14.52} \\
    \bottomrule
  \end{tabular}
\end{table}

\subsection{Ablation Study}
We ablate the key designs in Tab.~\ref{tab:ablation}, with all variants trained on the
same data and number of steps.

\noindent\textbf{Pixel-aligned 3D visual encoding is essential for 3D grounding.}
Removing it (ID A \vs E) reduces Robo3D-P by 30.5 points and increases the Robo3D-T
RMSE by 6.17, while W2P drops by only 1.2 points: without geometry aligned with the
visual tokens, the model still locates targets in the image plane but fails to infer
their 3D positions.

\noindent\textbf{RFT generalizes embodied reasoning to 3D targets.} With the same data,
RFT (ID C) improves Robo3D-P over SFT (ID B) by 22.2 points, more than twice its
9.7-point gain on W2P, as it reinforces any prediction satisfying the 3D constraint
rather than imitating a single answer. Thinking further improves Robo3D-P by 4.6
points and reduces the Robo3D-T DFD by 1.09 (ID C \vs E).

\noindent\textbf{Rewards must be computed in 3D.} With image-plane rewards (ID D), W2P
remains almost unchanged (74.6\% \vs 75.0\%), but Robo3D-P drops by 19.4 points and
the Robo3D-T RMSE increases by 3.20, since such rewards credit predictions with correct
projections but wrong depths, reintroducing the pixel-to-space gap during training.
\section{Conclusion}
\label{sec:conclusion}
We present Robo3D-VLM, a 3D-aware VLM that grounds the full chain of embodied reasoning
in 3D, from spatial understanding, object referring, and functional affordance to
placement and object-centric and end-effector trajectories. It perceives metric
geometry through a pixel-aligned 3D visual encoding, acquires 3D spatial understanding
via SFT, and extends it to embodied reasoning via RFT with rewards computed in 3D.
Together with Robo3D-3M and Robo3D-Bench, extensive experiments show that Robo3D-VLM
outperforms existing methods on spatial understanding and embodied reasoning, and
achieves zero-shot manipulation in both simulation and the real world.

\section*{LLM Usage Statement}
Large language models were used to assist with language editing, LaTeX formatting,
and the preparation of figure layouts. All research ideas, methods, experiments,
and results were conceived, conducted, and verified by the authors, who take full
responsibility for the content of this paper.

{
    \small
    \bibliographystyle{ieeenat_fullname}
    \bibliography{main}
}


\end{document}